\documentclass[11pt,a4paper]{article}
\usepackage[hyperref]{emnlp2018}
\usepackage{times}
\usepackage{latexsym}

\usepackage{url}
 \usepackage{multirow}
\usepackage{tikz}
\usetikzlibrary{calc, positioning, arrows, automata,fit,shapes.geometric,backgrounds,shapes.misc}
\usepackage{tikz-dependency}
\usepackage{pgfplots}
\pgfplotsset{compat=1.12}
\usepackage{amsthm}
\usepackage{framed}
\usepackage{caption}
\usepackage{subcaption}
\usepackage{graphicx}
\usepackage{xcolor,colortbl}
\usepackage{algorithm2e}
\usepackage{amsmath}

\aclfinalcopy 

\mathchardef\mhyphen="2D

\newcommand{\squishlist}{
 \begin{list}{--}
  { \setlength{\itemsep}{0pt}
     \setlength{\parsep}{3pt}
     \setlength{\topsep}{3pt}
     \setlength{\partopsep}{0pt}
     \setlength{\leftmargin}{1.5em}
     \setlength{\labelwidth}{1em}
     \setlength{\labelsep}{0.5em} } }

\newcommand{\squishend}{
  \end{list}  }

\definecolor{Gray}{gray}{0.85}
\definecolor{lGray}{gray}{0.95}
\newcommand{\GG}[1]{}
  \newcolumntype{a}{>{\columncolor{lGray}}r}
\newcolumntype{b}{>{\columncolor{lGray}}r}

\title{Using Linguistic Features to Improve the Generalization Capability \\of Neural Coreference Resolvers}

\author{Nafise Sadat Moosavi$^{1,2}$\thanks{\hspace{0.7em}This author is currently employed by the Ubiquitous Knowledge Processing (UKP) Lab, Technische Universit\"at Darmstadt,  https://www.ukp.tu-darmstadt.de.} \and Michael Strube$^{1}$
       \\
       $^1$Heidelberg Institute for Theoretical Studies gGmbH\\
       $^2$Research Training Group AIPHES\\
	   \small \url{moosavi@ukp.informatik.tu-darmstadt.de}, \url{michael.strube@h-its.org}}

\date{}

\begin{document}
\maketitle
\begin{abstract}
Coreference resolution is an intermediate step for text understanding.
It is used in tasks and domains for which we do not necessarily have coreference annotated corpora.
Therefore, generalization is of special importance for coreference resolution.
However, while recent coreference resolvers have notable improvements on the CoNLL dataset,
they struggle to generalize properly to new domains or datasets.
In this paper, we investigate the role of linguistic features in building more generalizable coreference resolvers.
We show that generalization improves only slightly by merely using a set of additional linguistic features.
However, employing features and subsets of their values that are informative for coreference resolution,
considerably improves generalization.
Thanks to better generalization, our system
achieves state-of-the-art results in out-of-domain evaluations,
e.g., on WikiCoref, 
our system, which is trained on CoNLL, achieves on-par performance with a system designed for this dataset.

\end{abstract}

\section{Introduction}
Coreference resolution is the task of recognizing different expressions that refer to the same entity.
The referring expressions are called mentions.
For instance, the sentence ``[Susan]$_1$ sent [her]$_1$ daughter to a boarding school'' contains two coreferring mentions. ``her'' is an anaphor which refers to the antecedent ``Susan''.

The availability of coreference information benefits various Natural Language Processing (NLP) tasks including 
automatic summarization, question answering, machine translation and information extraction. 
Current coreference developments are almost only targeted at improving scores on the CoNLL official test set.
However,  
the superiority of a coreference resolver on the CoNLL evaluation sets does not necessarily indicate
that it also performs better on new datasets.
For instance, the ranking model of \newcite{clarkkevin16a}, the reinforcement learning model of \newcite{clarkkevin16b} 
and the end-to-end model of \newcite{leekenton17} are three recent coreference resolvers,
among which the model of \newcite{leekenton17} performs the best and that of \newcite{clarkkevin16b} 
performs the second best on the CoNLL development and test sets.
However, if we evaluate these systems on the WikiCoref dataset \cite{ghaddar16a}, 
which is consistent with CoNLL with regard to coreference definition and annotation scheme, 
the performance ranking would be in a reverse order\footnote{The single model of \newcite{leekenton17} is used here.}.

In \newcite{moosavi17b}, we investigate the generalization problem in coreference resolution
and show that there is a large overlap between the coreferring mentions in the CoNLL training and evaluation sets.
Therefore, higher scores on the CoNLL evaluation sets do not necessarily indicate a better coreference model. They
may be due to better memorization of the training data.
As a result, despite the remarkable improvements in coreference resolution,
the use of coreference resolution in other applications is mainly limited to the use of simple rule-based 
systems, e.g.\ \newcite{lapata05a},\newcite{yu2016unsupervised}, and \newcite{elsner08b}.

In this paper, we explore the role of linguistic features for improving generalization.
The incorporation of linguistic features is considered as a potential solution for building more generalizable NLP 
systems\footnote{E.g.\ there is a dedicated workshop for this topic \url{https://sites.google.com/view/relsnnlp}.}.
While linguistic features\footnote{We refer to features that are based on linguistic intuitions, e.g.\ string match, 
or are acquired from linguistic preprocessing modules, e.g.\ POS tags, as linguistic features.} were shown to be important for coreference resolution, e.g.\ \newcite{uryupina07} and \newcite{bengtson08},
state-of-the-art systems no longer use them and mainly rely on word embeddings and deep neural networks. 
Since all recent systems are using neural networks, we focus on the effect of linguistic features on a neural coreference resolver. 

The contributions of this paper are as follows: 

\squishlist
\item We show that linguistic features are more beneficial for a neural coreference resolver if we incorporate 
features and subsets of their values that are informative for discriminating coreference relations.
Otherwise, employing linguistic features with all their values  
only slightly affects the performance and generalization.
\item We propose an efficient discriminative pattern mining algorithm, called EPM, for determining (feature, value) pairs that are informative for the given task.
We show that while the informativeness of EPM mined patterns is on-par with those of its counterparts,
it scales best to large datasets.\footnote{The EPM code is available at \url{https://github.com/ns-moosavi/epm}}  
\item By improving generalization, 
we achieve state-of-the-art performance on all examined out-of-domain evaluations. 
Our out-of-domain performance on WikiCoref is on-par with that of \newcite{ghaddar16b}'s coreference resolver, which is a system specifically designed for WikiCoref and uses its domain knowledge.
\squishend
\section{Importance of Features in Coreference}
\label{ch:improvements:related}
{\newcite{uryupina07}}'s thesis is one of the most thorough analyses of linguistically motivated features for coreference resolution.
She examines a large set of linguistic features, 
i.e.\ string match, syntactic knowledge, semantic compatibility, discourse structure and salience, 
and investigates their interaction with coreference relations.
She shows that even imperfect linguistic features, which are extracted using error-prone preprocessing modules, 
boost the performance
and argues that coreference resolvers could and should benefit from linguistic theories.
Her claims are based on analyses on the MUC dataset.
\newcite{ng02a}, \newcite{yang04}, \newcite{ponzetto06b}, \newcite{bengtson08}, and \newcite{recasens09} also study the importance of features in coreference resolution.

Apart from the mentioned studies, which are mainly about the importance of individual features,
studies like \newcite{bjoerkelund12}, \newcite{fernandes12}, and \newcite{uryupina15} generate new features by combining basic features.
{\newcite{bjoerkelund12}} do not use a systematic approach for combining features.
{\newcite{fernandes12}} use the Entropy guided Feature Induction (EFI) approach \cite{fernandes12entropy}
to automatically generate discriminative feature combinations.
The first step is to train a decision tree on a dataset in which
each sample consists of features describing a mention pair. 
The EFI approach traverses the tree from the root in a depth-first order 
and recursively builds feature combinations. 
Each pattern that is generated by EFI starts from the root node.
As a result, EFI tends to generate long patterns.
A decision tree does not represent all patterns of data.
Therefore, it is not possible to explore all feature combinations from a decision tree.

{\newcite{uryupina15}} propose an alternative approach to EFI.
They formulate the problem of generating feature combinations as a pattern mining approach.
They use the Jaccard Item Mining (JIM) algorithm\footnote{\url{http://www.borgelt.net/jim.html}} \cite{segond11}.
They show that the classifier that uses the JIM features significantly outperforms the one that employs the EFI features.
\section{Baseline Coreference Resolver}
deep-coref \cite{clarkkevin16a} and e2e-coref \cite{leekenton17} are among the best performing coreference resolvers 
from which e2e-coref performs better on the CoNLL test set.
deep-coref is a pipelined system, i.e.\ a mention detection first determines the list of candidate mentions with their corresponding features.
It contains various coreference models including the mention-pair, mention-ranking, and entity-based models.
The mention-ranking model of deep-coref has three variations: (1) ``ranking'' uses the slack-rescaled max-margin training objective of \newcite{wiseman15},
(2) ``reinforce'' is a variation of the ``ranking'' model in which the hyper-parameters are set in a reinforcement learning framework \cite{sutton1998reinforcement},
and (3) ``top-pairs'' is a simple variation of the ``ranking'' model that uses a probabilistic objective function and is used for pretraining the ``ranking'' model. 

e2e-coref is an end-to-end system that jointly models mention detection and coreference resolution.
It considers all possible (start, end) word spans of each sentence as candidate mentions.
Apart from a single model, e2e-coref includes an ensemble of five models.

We use deep-coref as the baseline in our experiments.
The reason is that 
some of the examined features require the head of each mention to be known, e.g. head match, 
while e2e-coref mentions do not have specific heads and heads are automatically determined using an attention mechanism.
We also observe that if we limit e2e-coref candidate spans to those that correspond to deep-coref's detected mentions,
the performance of e2e-coref drops to a level on-par with deep-coref\footnote{
The CoNLL score of the e2e-coref single model on the CoNLL development set drops from $67.36$ to $65.81$, 
while that of the deep-coref ``ranking'' model is $66.09$.}.
\section{Examined Features}
\label{sect:base_features}
The examined linguistic features include string match, syntactic, shallow semantic and discourse features.
\textbf{Mention-based} features include:
\squishlist
\item Mention type: proper, nominal or pronominal
\item Fine mention type: proper, definite or indefinite nominal, or the citation form of pronouns
\item Gender: female, male, neutral, unknown
\item Number: singular, plural, unknown
\item Animacy: animate, inanimate, unknown
\item Named entity type: person, location, organization, date, time, number, etc.
\item Dependency relation: enhanced dependency relation \cite{SCHUSTER16.779} of the head word to its parent
\item POS tags of the first, last, head, two words preceding and following of each mention
\squishend

\textbf{Pairwise} features include:
\squishlist
\item Head match: both mentions have the same head, e.g.\ ``red hat'' and ``the hat''
\item String of one mention is contained in the other, e.g.\ ``Mary's hat'' and ``Mary''
\item Head of one mention is contained in the other, e.g.\ ``Mary's hat'' and ``hat''
\item Acronym, e.g.\ ``Heidelberg Institute for Theoretical Studies'' and ``HITS'' 
\item Compatible pre-modifiers: the set of pre-modifiers of one mention is contained in that of the other, e.g.\ ``the red hat that she is wearing'' and ``the red hat''
\item Compatible\footnote{One value is unknown, or both values are identical.}\ gender,\ e.g.\ ``Mary'' and ``women''
\item Compatible number, e.g.\ ``Mary'' and ``John''
\item Compatible animacy, e.g.\ ``those hats'' and ``it'' 
\item Compatible attributes: compatible gender, number and animacy, e.g.\ ``Mary'' and ``she'' 
\item Closest antecedent that has the same head and compatible premodifiers, e.g.\ ``this new book'' and ``This book'' in ``Take a look at this new book. This book is one of the best sellers.''
\item Closest antecedent that has compatible attributes, e.g.\ the antecedent ``Mary'' and the anaphor ``she'' in the sentence ``John saw Mary, and she was in a hurry''
\item Closest antecedent that has compatible attributes and is a subject, e.g.\ the antecedent ``Mary'' and the anaphor ``she'' in the sentence ``Mary saw John, but she was in a hurry'' 
\item Closest antecedent that has compatible attributes and is an object, e.g.\ ``Mary'' and ``she'' in ``John saw Mary, and she was in a hurry''
\squishend
The last three features are similar to the discourse-level features discussed by \newcite{uryupina07},
which are created by combining \emph{proximity}, \emph{agreement} and \emph{salience} properties.
She shows that such features are useful for resolving pronouns.
we estimate proximity by considering the distance of two mentions.
The salience is also incorporated by discriminating subject or object antecedents.  
We do not use any gold information. All features are extracted using Stanford CoreNLP \cite{corenlp}.

\section{Impact of Linguistic Features}
\label{sect:all}
In this section, we examine the effect of employing all linguistic features described in Section~\ref{sect:base_features} in a neural coreference resolver, i.e.\ deep-coref.
We use \emph{MUC} \cite{vilain95}, \emph{B}$^3$ \cite{bagga98b},
\emph{CEAF}$_e$ \cite{luoxiaoqiang05a}, \emph{LEA} \cite{moosavi16b},
and the \emph{CoNLL} score \cite{pradhan14}, i.e.\ the average F$_1$ value of \emph{MUC}, \emph{B}$^3$, and \emph{CEAF}$_e$, for evaluations.

The results of employing those features in deep-coref's ``ranking'' and ``top-pairs'' models on the CoNLL development set are reported in Table~\ref{tab:dev-linguistics}.

\begin{table}[htbp]
    \begin{center}\footnotesize
    \resizebox{\columnwidth}{!}{%
    \begin{tabular}{@{}l|@{\hskip3pt}r@{\hskip3pt}|@{\hskip3pt}r@{\hskip3pt}|@{\hskip3pt}r@{\hskip3pt}|@{\hskip3pt}r@{\hskip3pt}||@{\hskip3pt}r@{\hskip3pt}}
     \multicolumn{1}{c}{} & \multicolumn{1}{c}{MUC} &
     \multicolumn{1}{c}{$B^3$} & \multicolumn{1}{c}{CEAF$_e$} & \multicolumn{1}{c}{CoNLL} & \multicolumn{1}{c}{LEA} \\ \hline
     \hline
     ranking  & $74.31$  & $64.23$  & $59.73$ & $66.09$ & $60.47$  \\
	+linguistic  & $74.35$  & $63.96$  & $60.19$ & $66.17$ & $60.20$  \\
     \hline
     top-pairs  & $73.95$ & $63.98$ & $59.52$  & $65.82$ & $60.07$ \\ 
     +linguistic & $74.32$ & $64.45$ & $60.19$ & $66.32$ & $60.62$ \\
    \end{tabular}
    }
    \end{center}
    \caption{Impact of linguistic features on deep-coref models on the CoNLL development set.
	}
    \label{tab:dev-linguistics}
\end{table}
The rows ``ranking'' and ``top-pairs'' show the base results of deep-coref's ``ranking'' and ``top-pairs'' models, respectively. 
``+linguistic'' rows represents the results for each of the mention-ranking models in which the feature set of Section~\ref{sect:base_features} is employed.
The gender, number, animacy and mention type features, which have less than five values, 
are converted to binary features. 
Named entity and POS tags, and dependency relations
are represented as learned embeddings.

We observe that incorporating all the linguistic features bridges the gap between the performance of 
``top-pairs'' and ``ranking''.
However, it does not improve significantly over ``ranking''.
Henceforth, 
we use the ``top-pairs'' model of deep-coref as the baseline model to incorporate linguistic features.

To assess the impact on generalization, 
we evaluate ``top-pairs'' and ``+linguistic''\footnote{i.e.\ ``top-pairs+linguistic''} models that are trained on CoNLL,
on WikiCoref (see Table~\ref{tab:out_of_domain_linguistics}).
We observe that
the impact on generalization is also not notable, i.e.\
the CoNLL score improves only by 0.5pp over ``ranking''.
  
\begin{table}[htbp]
    \begin{center}\footnotesize
    \resizebox{\columnwidth}{!}{%
    \begin{tabular}{@{}l|@{\hskip3pt}r@{\hskip3pt}|@{\hskip3pt}r@{\hskip3pt}|@{\hskip3pt}r@{\hskip3pt}|@{\hskip3pt}r@{\hskip3pt}||@{\hskip3pt}r@{\hskip3pt}}
     \multicolumn{1}{c}{} & \multicolumn{1}{c}{MUC} &
     \multicolumn{1}{c}{$B^3$} & \multicolumn{1}{c}{CEAF$_e$} & \multicolumn{1}{c}{CoNLL} & \multicolumn{1}{c}{LEA} \\ \hline
	 ranking & $63.10$ & $48.43$ &  $47.18$ & $52.90$ & $44.40$  \\ 
	 top-pairs  & $63.09$ & $48.42$ & $46.05$ & $52.52$ &  $44.21$\\
	 +linguistic & $63.99$ & $49.63$ & $46.60$ & $53.40$ & $45.66$\\
    \end{tabular}
    }
    \end{center}
    \caption{Out-of-domain evaluation of deep-coref models on the WikiCoref dataset.
	}
    \label{tab:out_of_domain_linguistics}
\end{table}

Based on an ablation study, while our feature set contains numerous features, the resulting improvements of ``linguistic'' over ``top-pairs'' mainly comes from the last four pairwise features in Section~\ref{sect:base_features},
which are carefully designed features.

\section{Better Exploiting Linguistic Features}
\label{ch:improvements:our_app}
As discussed by \newcite{moosavi17b},
there is a large lexical overlap between the coreferring mentions
of the CoNLL training and evaluation sets.
As a result, lexical features provide a very strong signal for resolving coreference relations.

For linguistic features to be more effective in current coreference resolvers,
which rely heavily on lexical features,
they should also provide a strong signal for coreference resolution.

Additional linguistic features are not necessarily all informative for coreference resolution, especially if they are extracted automatically and are noisy.
Besides, for features with multiple values, e.g.\ mention-based features, 
only a small subset of values may be informative.

To better exploit linguistic features, 
we only employ (feature, value) pairs\footnote{Henceforth, we refer to them as feature-values.} 
that are informative for coreference resolution.
Coreference resolution is a complex task in which features have complex interactions \cite{recasens09}.
As a result, we cannot determine the informativeness of feature-values in isolation.

We use a discriminative pattern mining approach \cite{cheng07,cheng08,batal10}
that examines all combinations of feature-values, up to a certain length, 
and determines which feature-values are informative when they are considered in combination.

Due to the large data size (all mention-pairs of the CoNLL training data)
and the high dimensionality of 
feature-values, compared to common evaluation sets of pattern mining methods, 
the existing discriminative pattern mining approaches were not applicable to our data.
In this section, we propose an efficient discriminative pattern mining approach, called Efficient Pattern Miner (EPM), that is scalable to large NLP datasets. 
The most important properties of EPM are (1) it examines all frequent feature-values combinations, up to the desired length, 
(2) it is scalable to large datasets, and (3) it is only data dependent and independent of the coreference resolver.
\subsection{Notation}
\label{sect:back}
We use the following notations and definitions throughout this section:
\squishlist
\item $D=\{X_i,c(X_i)\}_{i=1}^n$: set of $n$ training samples. 
$X_i$ is the set of feature-values that describes the $i$th sample. 
$c(X_i) \in C$ is the label of $X_i$, e.g.\ coreferent and non-coreferent.
\item $A=\{a_1,\dots,a_l\}$: set of all feature-values present in $D$.
Each $a_i \in A$ is called an item, e.g.\ $a_i=$``anaphor type=proper''.
\item $p$: pattern $p=\{a_{i_1},\dots, a_{i_k}\}$ is a set of one or more items, e.g.\ $p=$\{``anaphor type=proper'', ``antecedent type=proper''\}. 
\item $support(p,c_i)$: the number of samples that contain pattern $p$ and are labeled with $c_i$. 
\squishend

\subsection{Data Structure}
\label{fptree}
For representing the input samples, we use the Frequent Pattern Tree (FP-Tree) structure that is the data structure of the FP-Growth algorithm \cite{han04}, i.e.\ one of the most common algorithms for frequent pattern mining.
FP-Tree provides a structure for representing all existing patterns of data in a compressed form.
Using the FP-Tree structure allows an efficient enumeration of all frequent patterns.
In the FP-Tree structure, items are arranged in descending order of frequency.
Frequency of an item corresponds to $\sum_{c_i \in C} \: support(a_i, c_i)$.
Except for the root,
which is a null node,
each node $n$ contains an item $a_i \in A$.
It also contains the support values of $a_i$
in the subpath of the tree that starts from the root and ends with $n$, i.e.\ $support_n(a_i,c_j)$.  

The FP-Tree construction method \cite{han04} is as follows:
(a) scan $D$ to collect the set of all items, i.e.\ $A$. 
 Compute $support(a_i,c_j)$ for each item $a_i \in A$ and label $c_j \in C$.
Sort $A$'s members in descending order according to their frequencies, i.e. $\sum_{c_i \in C} \: support(a_i, c_i)$. 
(b) create a null-labeled node as the root, and (c) scan $D$ again. For each $(X_i, c(X_i)) \in D$: 
\begin{enumerate}
 \item Order all items $a_j \in X_i$ according to the order in $A$.
 \item Set the current node ($T$) to the root. 
\item \label{prevstep} Consider $X_{i}=[a_k|\bar {X_i}]$, where $a_k$ is the first (ordered) item of $x_{_i}$, 
and $\bar {X_i}=X_i-a_k$.
If $T$ has a child $n$ that contains $a_k$ 
then increment $support_n(a_k,c(X_i))$ by one.
Otherwise, create a new node $n$ that contains $a_k$ with $support_n(a_k,c(X_i))=1$. 
Add $n$ to the tree as a child of $T$. 
\item If $\bar {X_i}$ is non-empty, set $T$ to $n$. 
 Assign $X_i = \bar {X_i}$ and go to step~\ref{prevstep}.  
\end{enumerate}

As an example, assume $D$ contains the following two samples:
\begin{itemize}
 \item [] $X_1$=\{ana-type=NAM, ant-type=NAM, head-match=F\}, $C(X_1)=0$
 \item[]  $X_2$=\{ana-type=NAM, ant-type=NAM, head-match=T\}, $C(X_2)=1$
\end{itemize}
Based on these samples $A$=\{ana-type=NAM, ant-type=NAM, head-match=F, head-match=T\}, $support(a_i,0)_{a_i \in A}$= \{1,1,1,0\}, and $support(a_i,1)_{a_i \in A}$=\{1,1,0,1\}.
If we sort $A$ based on $a_i$'s frequencies ($support(a_i,0)+support(a_i,1)$), the ordering of $A$'s items will remain the same.

\begin{figure}[!htb]
\minipage{0.35\columnwidth}
\begin{tikzpicture}
	\footnotesize
\node [rectangle,draw]{ROOT} [level distance=10mm,sibling distance=30mm]
child { node [rectangle,draw]{ana=NAM (1,0)} [level distance=10mm ,sibling distance=15mm]
child {node [rectangle,draw] {ant=NAM (1,0)}
child {node [rectangle,draw,minimum width=22mm] {head=F (1,0)}
}}};
\end{tikzpicture}
\endminipage\hfill
\minipage{0.65\columnwidth}
\begin{tikzpicture}
	\footnotesize
	
\node [rectangle,draw]{ROOT} [level distance=10mm,sibling distance=30mm]
child { node [rectangle,draw]{ana=NAM (1,1)} [level distance=10mm ,sibling distance=30mm]
child {node [rectangle,draw] {ant=NAM (1,1)}
child {node [rectangle,draw] {head=F (1,0)}}
child {node [rectangle,draw,minimum width=5mm] {head=T (0,1)}}
}};
\end{tikzpicture}
\endminipage
\caption[FP-Tree construction steps]{Left to right: (partially) constructed FP-Tree for the example in Section~\ref{fptree}.\label{fp-tree_example}}
\end{figure}
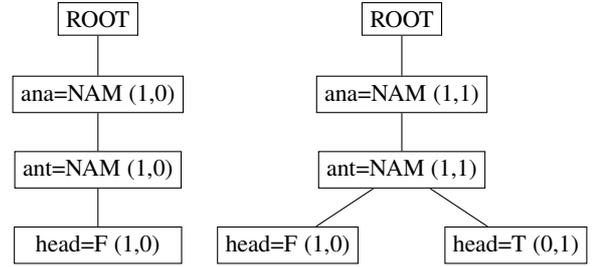

The FP-Tree construction steps for the above samples are demonstrated in Figure~\ref{fp-tree_example}.
ana-type, ant-type, and head-match features are abbreviated as
ana, ant, and head, respectively.

From an initial FP-Tree ($T$) that represents all existing patterns, 
one can easily obtain a new FP-Tree in which all patterns include a given pattern $p$.
This can be done by only including sub-paths of $T$ that contain pattern $p$. 
The new tree is called conditional FP-Tree of $p$, $T_{p}$.
An example of conditional FP-Tree is included in the supplementary materials.
%
\subsection{Informativeness Measures}
\label{patternEval}
We use a \emph{discriminative power} and an \emph{information novelty} measure for determining informativeness.
We also use a \emph{frequency} measure which determines the required minimum frequency of a pattern in training samples.
It helps to avoid overfitting to the properties of the training data.

%
 \noindent\textbf{Discriminative power}: We use the $G^2$ likelihood ratio test \cite{agresti07} in order to 
choose patterns whose association with the class variable 
is statistically significant.\footnote{A pattern is considered discriminative if 
the corresponding p-value is less than a fixed threshold ($0.01$).}
The $G^2$ test is successfully used for text analysis \cite{dunning93}.
%

 \noindent \textbf{Information Novelty}:
A large number of redundant patterns 
can be generated by adding irrelevant items to a base pattern that is discriminative itself.
We consider the pattern $p$ as novel if (1) $p$ predicts the target class label $c$ significantly
better than all of its containing items, and 
(2) $p$ predicts $c$ significantly better than all of its sub-patterns that 
satisfy the frequency, discriminative power,  
and the first information novelty conditions.
Similar to \newcite{batal10}, we employ 
a binomial distribution to determine information novelty.
\subsection{Mining Algorithm}
\label{mining}
%
The EPM algorithm is summarized in Algorithm~\ref{mine_patt}.
It takes FP-Tree $T$, pattern $p$ on which $T$ is conditioned, 
and set of items ($A_j \subset A$) whose combinations with $p$ will be examined. 
Initially, $p$ is empty and the FP-Tree is constructed based on all frequent items of data
and $A_j = A$. Resulting patterns are collected in $P$.

For each $a_i \in A_j$, the algorithm builds new pattern $q$ by combining $a_i$ with $p$.
$frequent(q)$ checks whether $q$ meets the frequency condition.
If $q$ is frequent, the algorithm continues the search process.
Otherwise, it is not possible to build any frequent pattern out of a non-frequent one.
\emph{Discriminative power} and the first condition of \emph{information novelty} are then checked for pattern $q$.
\begin{algorithm}[htp]
\SetAlgoLined
\DontPrintSemicolon 
\SetAlFnt{\small}
\SetKwFunction{algo}{$EPM$}
\SetKwProg{myalg}{Algorithm}{}{}
  \myalg{\algo{$T$, $p$, $A_j$}}{
  \ForEach {$a_i \in A_j$}{
  $q = p \cup \{a_i$\} \;
  \If{$Frequent(q)$}{
      \If{$Discriminative(q)$}{
	  \If{$Novel(q)$}{
	      $P = P \cup q$ \;
	  }
      }
      
      \If{$|q| >= \Theta_l$}{
	continue\;
      }
      \text{construct $T_{q}$ = $q$'s conditional tree} \;
      \vspace*{-.4cm}
       $EPM(T_{q}, q, ancestors(a_i))$ \;
   }
   }
   \vspace*{0.1cm}
}
\caption{The EPM algorithm.}
\label{mine_patt}
\end{algorithm}

We use a threshold ($\Theta_l$) 
for the maximum length of mined patterns.
$\Theta_l$ can be set to large values if more complex and specific patterns are desirable.

If $|q|$ is smaller than $\Theta_l$,
the conditional FP-Tree $T_{q}$ is built that represents patterns of $T$ that include the pattern $q$.
The mining algorithm then continues 
to recursively search for more specific patterns by combining $q$ with the items included in $ancestors(a_i)$,
which keeps the list of all ancestors of $a_i$ in the original FP-Tree.
EPM examines all frequent patterns of up to length $\Theta_l$. 
%

If we use a statistical test multiple times, the risk of making false discoveries increases \cite{webb2006}.
To tackle this, we apply the Bonferroni correction for multiple tests in a post-pruning function after the mining process.
This function also applies the second information novelty condition on the resulting patterns.
\begin{table*}[!htb]
\begin{center}\footnotesize
    \resizebox{\textwidth}{!}{%
  \begin{tabular}{@{}l||l|r||r||r|r||r|r|r|r|r||r|r|r|r@{}}
  \hline
  \multicolumn{1}{c||}{} & \multicolumn{3}{c||}{Data characteristics} & \multicolumn{3}{c||}{\# Patterns} & \multicolumn{4}{c||}{Micro-F} & \multicolumn{4}{c}{Macro-F} \\ \hline \hline
  \multicolumn{1}{c||}{Dataset} &\multicolumn{1}{c|}{\#Features} & \multicolumn{1}{c|}{\#FI} & \multicolumn{1}{c||}{$n$} & DDP & MPP & EPM & Orig & DDP & MPP & EPM & Orig & DDP & MPP &EPM \\ \hline
  \hline
  \small cmc & $(0/2/7)$ & $24$ & $1473$ & $4$ & $99$ & $23$ & $77.5$ & $77.4$ & $76.2$ & $77.3$ & $57.3$ & $57.1$ & $57.7$ & $59.4$ \\
  \small nursery  & $(0/0/8)$ & $27$ & $12690$ & $4$ & $258$ & $198$ & $97.5$ & $98.2$ & $99.9$ & $99.8$ & $49.4$ & $79.4$ & $99.8$ & $98.8$\\
  \small sick & $(6/1/22)$ & $36$ & $2800$ & $5$ & $627$ & $89$ & $94.6$ & $94.7$ & $96.1$ & $95.8$ & $62.6$ & $64.8$ & $81.0$ & $75.6$\\
  \small kr-v-k & $(0/0/16)$ & $40$ & $28056$ & $7$ & $71$ & $63$ & $99.1$ & $99.1$ & $99.6$ & $99.6$ & $49.8$ & $49.8$ & $87.8$ & $88.4$ \\
  \small german & $(0/7/13)$ & $51$ & $1000$ & $8$ & $548$ & $97$ & $70.7$ & $70.9$ & $73.1$ & $72.7$ & $49.6$ & $55.2$ & $65.3$ & $64.2$\\ 
  \small connect-4 & $(0/0/42)$ & $75$ & $67557$ & - & - & $907$ & $90.5$ & - & - & $90.5$ & $47.5$ & - & - & $56.6$\\ 
  \small census & $(1/12/28)$ & $76$ & $299284$ & - & - & $5618$& $93.8$ & - & - & $93.8$ & $48.4$ & - & - & $51.6$ \\
  \small poker & $(0/10/0)$ & $85$ & $1025010$ & - & - & $14216$ & $23.1$ & - & - & $49.6$ & $22.4$ & - & - & $44.5$ \\ 
  \end{tabular}
 }
   \caption[Evaluating the informativeness of DDPMine, MPP and EPM patterns]{Evaluating the informativeness of DDPMine, MPP and EPM patterns on standard datasets.}
  \label{tab:performance}
 \end{center}
\end{table*}
\section{Why Use EPM?}
In this section, we explain why EPM is a better alternative compared to its counterparts for large NLP datasets.
We compare EPM with two efficient discriminative pattern mining algorithms, i.e.\
Minimal Predictive Patterns (MPP) \cite{batal10} and Direct Discriminative Pattern Mining (DDPMine) \cite{cheng08},
on standard machine learning datasets.

MPP selects patterns that are significantly more predictive than all their sub-patterns.
It measures significance by the binomial distribution.
For each pattern of length $l$, MPP checks $2^l-1$ sub-patterns.
DDPMine is an iterative approach that selects 
the most discriminative pattern at each iteration and reduces 
the search space of the next iteration by removing all samples that include 
the selected pattern. DDPMine uses the FP-Tree structure.

We show that EPM scales best and compares favorably based on the informativeness of resulting patterns.
Due to its efficiency, EPM can handle large datasets similar 
to ones that are commonly used in various NLP tasks.
\subsection{Experimental Setup}
We use the same FP-Tree implementation for DDPMine and EPM.
In all algorithms, we consider a pattern as frequent if it occurs in 10\% of the samples of one of the classes.
We use $\Theta_l=3$ for both MPP and EPM.

We perform 5-times repeated 5-fold cross validation and the results are averaged.
In each validation, all experiments are performed on the same split.
We use a linear SVM, i.e.\  LIBLINEAR 2.11 \cite{fan08}, as the baseline classifier.

We use several datasets from the UCI machine learning repository \cite{Lichman:2013} whose
characteristics are presented in the first three columns of Table~\ref{tab:performance},
i.e.\ the number of (1) (real/integer/nominal) features (\#Features), (2) frequent items (\#FI), and (3) samples ($n$). 
We use one[the minority class]-vs-all technique for datasets with more than two classes.
\subsection{How Informative are EPM Patterns?}
To evaluate the informativeness of mined patterns,
the common practice is to add them as new features to the feature set of the baseline classifier;
the more informative the patterns, the greater impact they would have on the overall performance.
All patterns are added as binary features, i.e.\ 
the feature is true for samples that contain all items of the corresponding pattern.

The effect of the patterns of DDPMine, MPP and EPM on the overall accuracy 
is presented in Table~\ref{tab:performance}.  
The columns \#Patterns show the number of patterns mined by each of the algorithms. 
The \emph{Orig} columns
show the results of the SVM using the original feature sets.
The \emph{DDP}, \emph{MPP}, and \emph{EPM} columns show the results of the SVM 
on the datasets for which the feature set is extended by 
the features mined by DDPMine, MPP, and EPM, respectively.
The results of the 5-repeated 5-fold cross validation are reported
if each single validation takes less than 10 hours. 

Based on the results of Table~\ref{tab:performance} (1) EPM efficiently scales to larger datasets,
(2) MPP and EPM patterns considerably improves the performance,
and (3) EPM has on-par results with MPP while it mines considerably fewer patterns.
\subsection{How Does it Scale?}
\begin{figure}[!htb]
\begin{center}
\begin{tikzpicture}[scale = .7]   %
\begin{axis}[
ymin = 0.01, ymax= 170000, legend entries = {EPM(3)\\EPM(4)\\ DDPMine\\ MPP(3)\\MPP(4)\\},ymode=log, 
legend columns=5, 
legend style={at={(1.2,-0.25)},font=\small}, legend cell align=center,
 draw=none,
 xtick=data,
 xticklabel = {\benchmark{\tick}},
 xtick align=outside,
 ytick align=inside,
 x tick label style={rotate=45,anchor=east},
 title style={font=\Large},
 ylabel near ticks,
 enlarge x limits=false,
xtick distance={400},
 symbolic x coords={cmc,nursery,sick,kr-v-k,german,connect,census,poker},
 ymajorgrids=true,
 grid = none,
 ]
\addplot[
    solid,
    color=black,
    mark=square,
    ]
    coordinates {
    (cmc, 0.001)(nursery, 1)(sick, 0.01)(kr-v-k, 1)(german, 0.01)(connect, 33)(census, 59)(poker, 84)
    };
    \addplot[
    dashed,
    color=black,
    mark=*,
    ]
    coordinates {
    (cmc, 0.001)(nursery, 1)(sick, 1)(kr-v-k, 1)(german,0)(connect, 185)(census, 85)(poker, 90)};
    \addplot[
    solid,
    color=red!60!black,
    mark=diamond,
    ]
    coordinates {
    (cmc, 0.001)(nursery, 10)(sick, 13663)(kr-v-k, 39)(german, 18)(connect, 172800)(census, 172800)(poker, 1090) };
    \addplot[
    color=blue,
    mark=square,
    ]
    coordinates {
    (cmc, 2.6)(nursery, 145)(sick, 2589)(kr-v-k, 120)(german, 186)(connect, 172800)(census, 172800)(poker,68794) };
    \addplot[
    dashed,
    color=blue,
    mark=*,
    ]
    coordinates {
    (cmc, 394)(nursery, 18515)(sick, 172800)(kr-v-k, 3833)(german,125514)(connect, 172800)(census, 172800)(poker, 172800) };

 \end{axis}
\end{tikzpicture}
\end{center}
\caption{Comparison of mining times (seconds).}
\label{fig:all}
\end{figure}
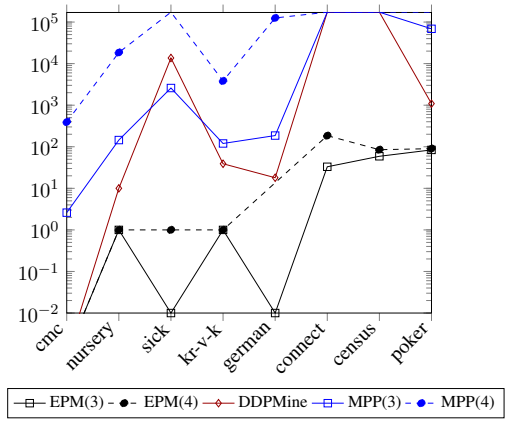
Figure~\ref{fig:all} compares EPM mining time (in seconds) with those of DDPMine and MPP.
The parameter in the parentheses is the pattern size threshold, e.g.\ $\Theta_l=4$ for EPM(4).
The experiments that take more than two days are terminated and are not included.
EPM is notably faster in comparison to the other two approaches.
It is notable that the examined datasets are considerably smaller than the coreference data, which includes more than 33 million samples and 200 frequent feature-values.
\begin{table*}[!htb]
    \begin{center}\footnotesize
    \resizebox{\textwidth}{!}{%
    \begin{tabular}{@{}l|l|@{\hskip3pt}rrr@{\hskip3pt}|@{\hskip3pt}rrr@{\hskip3pt}|@{\hskip3pt}rrr@{\hskip3pt}|r@{\hskip3pt}||@{\hskip3pt}rrr@{\hskip3pt}}
     \multicolumn{2}{c}{} & \multicolumn{3}{c}{MUC} &
     \multicolumn{3}{c}{$B^3$} & \multicolumn{3}{c}{CEAF$_e$} & \multicolumn{1}{c}{CoNLL} & \multicolumn{3}{c}{LEA} \\ \hline
     \multicolumn{1}{c}{}& \multicolumn{1}{c}{} &R & P & F$_1$ & R & P & F$_1$ & R & P & F$_1$ &  & R & P & F$_1$\\ \hline
     \hline
     \parbox[t]{2mm}{\multirow{5}{*}{\rotatebox[origin=c]{90}{ deep-coref}}} 
     & ranking & $70.43$ & $79.57$ & $74.72$ & $58.08$ & $69.26$ & $63.18$ & $54.43$ & $64.17$ & $58.90$ & $65.60$ & $54.55$ & $65.68$ & $59.60$ \\
     & reinforce & $69.84$ & $79.79$ & $74.48$ & $57.41$ & $70.96$ & $63.47$ & $55.63$ & $63.83$ & $59.45$ & $65.80$ & $53.78$ & $67.23$ & $59.76$\\ 
     &top-pairs & $69.41$ & $79.90$ & $74.29$ & $57.01$ & $70.80$ & $63.16$ & $54.43$ & $63.74$ & $58.72$ & $65.39$ & $53.31$ & $67.09$ & $59.41$  \\ 
     &{+EPM} & $71.16$ & $79.35$ & $75.03$ & $59.28$ & $69.70$ & $64.07$ & $56.52$ & $64.02$ & $60.04$ & $66.38$ & $55.63$ & $66.11$ & $60.42$ \\
     &+JIM & $69.89$ & $80.45$ & $74.80$ & $57.08$ & $71.58$ & $63.51$ & $55.36$ & $64.20$ & $59.45$ & $65.93$ & $53.46$ & $67.97$ & $59.85$\\
     \hline
     \parbox[t]{2mm}{\multirow{2}{*}{\rotatebox[origin=c]{90}{ e2e}}} 
     & single & $74.02$ & $77.82$ & $75.88$ & $62.58$ & $67.45$ & $64.92$ & $59.16$ & $62.96$ & $61.00$ & $67.27$ & $58.90$ & $63.79$ & $61.25$ \\
     & ensemble & $73.73$ & $80.95$ & $77.17$ & $61.83$ & $72.10$ & $66.57$ & $60.11$ & $65.62$ & $62.74$ & $68.83$ & $58.48$ & $68.81$ & $63.23$\\
     \hline
    \end{tabular}
    }
    \end{center}
    \caption{\footnotesize Comparisons on the CoNLL test set. 
    The F$_1$ gains that are statistically significant: (1) ``+EPM'' compared to ``top-pairs'', ``ranking'' and ``JIM'', (2) ``+EPM'' compared to ``reinforce'' based on MUC, B$^3$ and LEA, (3) ``single'' compared to ``+EPM'' based on MUC and B$^3$, and (4) 
``ensemble'' compared to other systems. Significance is measured based on the approximate randomization test ($p<0.05$) \cite{noreen89}.
    }
    \label{tab:in-domain-evaluations}
\end{table*}

\section{Impact of Informative Feature-values}
\subsection{Experimental Setup}
\label{ch:improvements_baseline}
For determining informative feature-values, 
we extract all features for all mention-pairs\footnote{Each mention is paired with all the preceding mentions.} of the CoNLL training data 
and then apply EPM on this 
data.
In order to prevent learning annotation errors and specific properties of the training data, 
we consider a pattern as frequent if it occurs in coreference relations of at least 
$m$ different coreferring anaphors ($m=20$).
Since the majority of mention-pairs are non-coreferent and we are not interested in patterns for non-coreferring relations, 
we also consider the coreference probability of each pattern $p$, i.e.\ $\frac{|\{X_i|p \in X_i \land c(X_i)=coreferent\}|}{|\{X_i|p \in X_i\}|}$, 
in the post-pruning function. 
The coreference probability should be higher than a threshold ($60\%$ in our experiments),
so we only mine patterns that are informative for coreferring mentions.

For the coreference resolution experiments, instead of incorporating informative patterns, we incorporate feature-values that are included in the informative patterns mined by EPM.
The reason
is that deep-coref, or any other recent coreference resolver, uses a deep neural network,
which has a fully automated feature generation 
process.
We add these feature-values 
as binary features.

By setting $\Theta_l$ to five,\footnote{We observe that using larger $\Theta_l$ values will result in many over-specified patterns.} EPM results in 
13 pairwise feature-values,
112 POS tags, i.e.\ 53 POS for anaphors and 59 for antecedents,
25 dependency relations,
26 mention types (mention types or fine mention types),
and finally, 14 named entity tags.\footnote{Following the previous studies that show different features are of different importance for various types of mentions, e.g.\ \newcite{denis08} and \newcite{moosavi17a},
we mine a separate set of patterns for each type of anaphor.
These resulting feature-values are the union of informative feature-values for all types of anaphora.}

Based on the observation in Section~\ref{sect:all}, we use the top-pairs model of deep-coref as the baseline to employ additional features, i.e. ``+EPM'' is the top-pairs model in which EPM feature-values are incorporated.
\subsection{Impact on In-domain Performance}
The performance of the ``+EPM'' model compared to recent state-of-the-art coreference models on the CoNLL test set is presented in Table~\ref{tab:in-domain-evaluations}.
The ``single'' and ``ensemble'' rows represent the results of the single and ensemble models of e2e-coref. 

We also compare EPM with the pattern mining approach used by \newcite{uryupina15}, i.e.\ Jaccard Item Mining (JIM).
For a fair comparison, while \newcite{uryupina15} used mined patterns for extracting feature templates, we use them for selecting feature-values.
We run the JIM algorithm on 
the same data and with the same setup as that of EPM.\footnote{
We set the minimum frequency, maximum pattern length and $score^+$ threshold parameters of JIM to 20, 5 and 0.6.}
This results in nine pairwise features, 
260 POS tags, 38 dependency relations, 32 mention types, and 18 named entity tags.
The ``+JIM'' row shows the results of deep-coref top-pairs model 
in which these feature-values are incorporated.
As we see, EPM feature-values result in significantly better performance than those of JIM
while the number of EPM feature-values is considerably less than JIM.
%
\begin{table}[!htb]
    \begin{center}\footnotesize
    \begin{tabular}{@{}l|r|r|r|r|r@{\hskip3pt}}
     \multicolumn{1}{c}{} & \multicolumn{1}{c}{MUC} & \multicolumn{1}{c}{$B^3$} & \multicolumn{1}{c}{CEAF$_e$} & \multicolumn{1}{c}{CoNLL} & \multicolumn{1}{c}{LEA} \\ \hline
     \hline
     +EPM  & $74.92$ & $65.03$ & $60.88$ & $66.95$ & $61.34$ \\ \hline
     -pairwise & $74.37$ & $64.55$ & $60.46$ & $66.46$ &$60.71$\\
     -type & $74.71$ & $64.87$ & $61.00$ & $66.86$ & $61.07$\\
     -dep & $74.57$ & $64.79$ & $60.65$ & $66.67$ & $61.01$\\
     -NER & $74.61$ & $65.05$ & $60.93$ & $66.86$ & $61.27$\\
     -POS & $74.74$ & $65.04$ & $60.88$ & $66.89$ & $61.30$\\ \hline
     +pairwise & $74.25$ & $64.33$ & $60.02$ & $66.20$ & $60.57$ \\ 
     \hline
    \end{tabular}
    \end{center}
    \caption{Impact of different EPM feature groups on the CoNLL development set.}
    \label{tab:feature-ablation}
\end{table}

\begin{table*}[!htb]
    \begin{center}\footnotesize
    \resizebox{\textwidth}{!}{%
    \begin{tabular}{@{}l|@{\hskip3pt}l|@{\hskip3pt}rrr@{\hskip3pt}|@{\hskip3pt}rrr@{\hskip3pt}|@{\hskip3pt}rrr@{\hskip3pt}|r@{\hskip3pt}||@{\hskip3pt}rrr@{\hskip3pt}}
     \multicolumn{2}{c}{} & \multicolumn{3}{c}{MUC} &
     \multicolumn{3}{c}{$B^3$} & \multicolumn{3}{c}{CEAF$_e$} & \multicolumn{1}{c}{CoNLL} & \multicolumn{3}{c}{LEA} \\ \hline
     \multicolumn{1}{c}{}& \multicolumn{1}{c}{} &R & P & F$_1$ & R & P & F$_1$ & R & P & F$_1$ &  & R & P & F$_1$\\ \hline
     \parbox[t]{2mm}{\multirow{5}{*}{\rotatebox[origin=c]{90}{ deep-coref}}} 
      &ranking & $57.72$ & $69.57$ & $63.10$ & $41.42$ & $58.30$ & $48.43$ & $42.20$ & $53.50$ & $47.18$ & $52.90$ & $37.57$ & $54.27$ & $44.40$  \\ 
      &reinforce & $62.12$ & $58.98$ & $60.51$ & $46.98$ & $45.79$ & $46.38$ & $44.28$ & $46.35$ & $45.29$ & $50.73$ & $42.28$ & $41.70$ & $41.98$ \\
      &top-pairs & $56.31$ & $71.74$ & $63.09$ & $39.78$ & $61.85$ & $48.42$ & $40.80$ & $52.85$ & $46.05$ & $52.52$ & $35.87$ & $57.58$ & $44.21$\\ 
      &{+EPM} & $58.23$ & $74.05$ & $\boldsymbol{65.20}$ & $43.33$ & $63.90$ & $51.64$ & $43.44$ & $56.33$ & $\boldsymbol{49.05}$ & $\boldsymbol{55.30}$ & $39.70$ & $59.81$ & $\boldsymbol{47.72}$ \\  
     \parbox[t]{2mm}{\multirow{2}{*}{\rotatebox[origin=c]{90}{ e2e}}} 
      & single & $60.14$ & $64.46$ & $62.22$ & $45.20$ & $51.75$ & $48.25$ & $38.18$ & $43.50$ & $40.67$ & $50.38$ & $40.70$ & $47.56$ & $43.86$ \\
      & ensemble & $59.58$ & $71.60$ & $65.04$ & $44.64$ & $60.91$ & $51.52$ & $40.38$ & $49.17$ & $44.35$ & $53.63$ & $40.73$ & $56.97$ & $47.50$\\ \hline
     & {G\&L} & $66.06$ & $62.93$ & $64.46$ & $57.73$ & $48.58$ & $\boldsymbol{52.76}$ & $46.76$ & $49.54$ & $48.11$ & $55.11$ & - & - & -\\ \hline
    \end{tabular}
    }
    \end{center}
    \caption{Out-of-domain evaluation on the WikiCoref dataset. The highest $F_1$ scores are boldfaced.}
    \label{tab:out-domain-deep-coref}
\end{table*}
\paragraph{Feature Ablation}
Table~\ref{tab:feature-ablation} shows the effect of each group of EPM feature-values, i.e.\ pairwise features, mention types, 
dependency relations, named entity tags and POS tags, on the performance of ``+EPM''.
The performance of ``+EPM'' from which each of the above feature groups is removed, one feature group at a time, 
is represented as ``-pairwise'', ``-types'', ``-dep'', ``-NER'', and ``-POS'', respectively.
The POS and named entity tags have the least 
and the pairwise features have the most significant effect.
Since pairwise features have the most significant effect, 
we also perform an experiment in which only pairwise features are incorporated in the ``top-pairs'' model, i.e.\ ``+pairwise''.
The results of ``-pairwise'' compared to ``+pairwise'' show that pairwise feature-values 
have a significant impact, but only when they are considered in combination with other EPM feature-values.
%

%
%
\begin{table}[htbp]
    \begin{center}\footnotesize
    \begin{tabular}{@{}l@{\hskip3pt}|@{\hskip3pt}l@{\hskip3pt}|@{\hskip3pt}r@{\hskip3pt}|@{\hskip3pt}r@{\hskip3pt}||@{\hskip3pt}r@{\hskip3pt}|@{\hskip3pt}r@{}}
         \multicolumn{2}{c}{} & \multicolumn{2}{c}{in-domain} & \multicolumn{2}{c}{out-of-domain} \\ \hline
	 \multicolumn{2}{c}{} & CoNLL & LEA & CoNLL & LEA \\ \hline
          \multicolumn{2}{c}{} & \multicolumn{4}{c}{{pt (Bible)} } \\ \hline
	 \multirow{2}{*}{{ deep-coref}}
     &ranking & $75.61$ & $71.00$ & $66.06$ & $57.58$  \\ 
     & +EPM & $76.08$ & $71.13$ & $\boldsymbol{68.14}$ & $\boldsymbol{60.74}$ \\ \hline
     \multirow{2}{*}{{ e2e-coref}} 
     & single& ${77.80}$ & ${73.73}$ & $65.22$ & $58.26$\\ 
	 & ensemble & $\boldsymbol{78.88}$ & $\boldsymbol{74.88}$ & $65.45$ & $59.71$\\
	 \hline 
     \hline
     \multicolumn{2}{c}{} & \multicolumn{4}{c}{{wb (weblog)} } \\ \hline
	 \multirow{2}{*}{{ deep-coref}}
     &ranking & $61.46$ & $53.75$ & $57.17$ & $48.74$ \\ 
     &+EPM & $61.97$ & ${53.93}$ & $\boldsymbol{61.52}$ & $\boldsymbol{53.78}$ \\ \hline
     \multirow{2}{*}{{ e2e-coref}}
     &single & ${62.02}$ & $53.09$ & $60.69$ & $52.69$\\ 
	 &ensemble & $\boldsymbol{64.76}$ & $\boldsymbol{57.54}$ & $60.99$ & $52.99$\\
	 
     \hline
    \end{tabular}
    \end{center}
    \caption{In-domain and out-of-domain evaluations for the pt and wb genres of the CoNLL test set.
    The highest scores are boldfaced.
    }
    \label{tab:cross_genre_enhanced_1}
\end{table}
\subsection{Impact on Generalization}
\label{ch:improvements:generalization}
We use the same setup as that of \newcite{moosavi17b} for evaluating generalization 
including (1) training on the CoNLL data and testing on WikiCoref\footnote{WikiCoref only contains 30 documents, which is not enough for training neural coreference resolvers.}
and (2) excluding a genre of the CoNLL data from training and development sets 
and testing on the excluded genre. Similar to \newcite{moosavi17b}, we use the \emph{pt} and \emph{wb} genres for the latter evaluation setup.  

The results of the first evaluation setup are shown in Table~\ref{tab:out-domain-deep-coref}.
The best performance on WikiCoref is achieved by \newcite{ghaddar16a} (``G\&L'' in Table~\ref{tab:out-domain-deep-coref})
who introduced WikiCoref and design a domain-specific coreference resolver that makes use of 
the Wikipedia markups of a document as well as links to Freebase, which are annotated in WikiCoref.

Incorporating EPM feature-values improves the performance by about three points.
While ``+EPM'' does not use the WikiCoref data during training, and unlike ``G\&L'', it does not employ any domain-specific features,
it achieves on-par performance with that of ``G\&L''.
This indeed shows the effectiveness of informative feature-values in improving generalization.



The second set of generalization experiments is reported in Table~\ref{tab:cross_genre_enhanced_1}. 
``in-domain'' columns show the results when the evaluation genres were included in training and development sets 
while the ``out-of-domain'' columns show the results when the evaluation genres were excluded.
As we can see, ``+EPM'' generalizes best, and in out-of-domain evaluations, it considerably outperforms the ensemble model of e2e-coref, which has the best performance on the CoNLL test set. 
\section{Conclusions}
In this paper, we show that employing linguistic features 
in a neural coreference resolver significantly improves generalization.
However, the incorporated features should be informative enough 
to be taken into account in the presence of lexical features, 
which are very strong features in the CoNLL dataset.
We propose an efficient algorithm to determine informative feature-values in large datasets.
As a result of a better generalization, we achieve state-of-the-art results in all examined out-of-domain evaluations.

\section*{Acknowledgments} The authors would like to thank Mark-Christoph M\"uller, Benjamin Heinzerling, Alex Judea, Steffen Eger and the anonymous reviewers for their helpful comments and feedbacks. 
This work has been supported by the Klaus Tschira Foundation, Heidelberg,
Germany and the German Research
Foundation (DFG) as part of the Research Training Group
“Adaptive  Preparation  of  Information  from  Heterogeneous  Sources”  (AIPHES) under  grant  No.
GRK 1994/1. 

\bibliographystyle{acl_natbib_nourl}
\bibliography{mybib}

\end{document}


\maketitle

\appendix
\section{Supplemental Material}
\label{sec:supplemental}

\subsection{Running Example for FP-Tree Construction}
Assume $D$ contains the following three samples:
\begin{itemize}
 \item [] $X_1$=\{ana-type=NAM, ant-type=NAM, head-match=F\}, $C(X_1)=0$
 \item[]  $X_2$=\{ana-type=NAM, ant-type=NAM, head-match=T\}, $C(X_2)=1$
 \item[]  $X_3$=\{ana-type=NAM, ant-type=NOM, head-match=F\}, $C(X_3)=0$
\end{itemize}
Based on these three samples
\begin{itemize}
\item $A$=\{ana-type=NAM, ant-type=NAM, head-match=F, head-match=T, ant-type=NOM\},
\item $support(a_i,0)_{a_i \in A}$= \{2,1,2,0,1\}, e.g.\ ``ana-type=NAM'' appeared two times in non-coreferring ($C(X_i)=0$) samples,
\item and $support(a_i,1)_{a_i \in A}$=\{1,1,0,1,0\}.
\end{itemize}
If we sort $A$ based on $a_i$'s frequencies, i.e. $support(a_i,0)+support(a_i,1)$, the ordering of $A$'s items will remain the same.

Now, we need to go through the samples again to build the tree.
The FP-Tree construction steps after adding each of the above samples is demonstrated in Figure~\ref{fp-tree_example}.
ana-type=NAM, ant-type=NAM, head-match=F, head-match=T, and ant-type=NOM are abbreviated as
ana=NAM, ant=NAM, head=F, head=T, and ant=NOM, respectively in Figures~\ref{fp-tree_example} and ~\ref{final-fp-tree_example}.
\begin{figure*}[!htb]
\minipage{0.28\textwidth}
\begin{tikzpicture}
\node [rectangle,draw]{ROOT} [level distance=10mm,sibling distance=30mm]
child { node [rectangle,draw]{ana=NAM} [level distance=10mm ,sibling distance=15mm]
child {node [rectangle,draw] {ant=NAM}
child {node [rectangle,draw] {head=F}
}}};
\end{tikzpicture}
\endminipage\hfill
\minipage{0.28\textwidth}
\begin{tikzpicture}
\node [rectangle,draw]{ROOT} [level distance=10mm,sibling distance=30mm]
child { node [rectangle,draw]{ana=NAM} [level distance=10mm ,sibling distance=20mm]
child {node [rectangle,draw] {ant=NAM}
child {node [rectangle,draw] {head=F}}
child {node [rectangle,draw] {head=T}}
}};
\end{tikzpicture}
\endminipage\hfill
\minipage{0.37\textwidth}%
\begin{tikzpicture}
\node [rectangle,draw]{ROOT} [level distance=10mm,sibling distance=25mm]
child { node [rectangle,draw]{ana=NAM} [level distance=10mm ,sibling distance=30mm]
child {node [rectangle,draw] {ant=NAM}  [level distance=10mm ,sibling distance=20mm]
child {node [rectangle,draw] {head=F}}
child {node [rectangle,draw] {head=T}}
}
child {node [rectangle,draw] {head=F}
child {node [rectangle,draw] {ant=NOM}
}}
};
\end{tikzpicture}
\endminipage
\caption[FP-Tree construction steps]{Left to right: (partial) constructed FP-Tree after adding each of the three given samples.
The right-most tree is the final FP-Tree that represents all input samples.\label{fp-tree_example}}
\end{figure*}
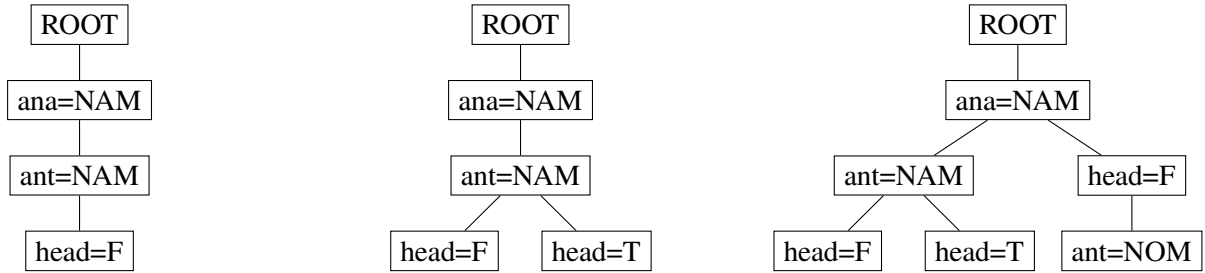
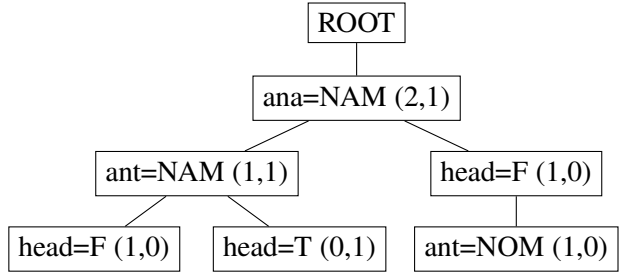
\begin{figure*}[!htb]
\begin{center}
\begin{tikzpicture}
\node [rectangle,draw]{ROOT} [level distance=10mm,sibling distance=25mm]
child { node [rectangle,draw]{ana=NAM (2,1)} [level distance=10mm ,sibling distance=42mm]
child {node [rectangle,draw] {ant=NAM (1,1)}  [level distance=10mm ,sibling distance=27mm]
child {node [rectangle,draw] {head=F (1,0)}}
child {node [rectangle,draw] {head=T (0,1)}}
}
child {node [rectangle,draw] {head=F (1,0)}
child {node [rectangle,draw] {ant=NOM (1,0)}
}}
};
\end{tikzpicture}
\end{center}
\caption[The final FP-Tree with support values]{FP-Tree with corresponding support values of the nodes.\label{final-fp-tree_example}}
\end{figure*}
Figure~\ref{final-fp-tree_example} shows the resulted FP-Tree in which corresponding support values for 
both classes, i.e.\ zero and one, are also included in each node.
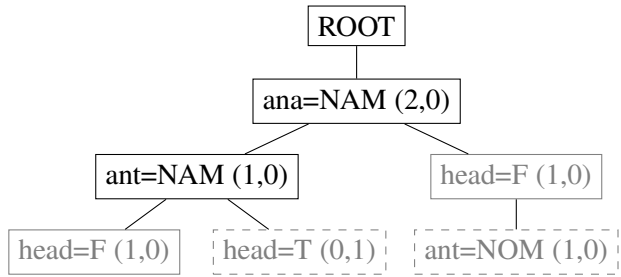
\begin{figure*}[!htb]
\begin{center}
\begin{tikzpicture}
\node [rectangle,draw]{ROOT} [level distance=10mm,sibling distance=25mm]
child { node [rectangle,draw]{ana=NAM (2,0)} [level distance=10mm ,sibling distance=42mm]
child {node [rectangle,draw] {ant=NAM (1,0)}  [level distance=10mm ,sibling distance=27mm]
child {node [rectangle,draw,gray] {head=F (1,0)}}
child {node [rectangle,draw,gray,dashed] {head=T (0,1)}}
}
child {node [rectangle,draw,gray] {head=F (1,0)}
child {node [rectangle,draw,gray,dashed] {ant=NOM (1,0)}
}}
};
\end{tikzpicture}
\end{center}
\caption{Conditional FP-Tree for the $p=\{\text{head=F}\}$ pattern.\label{conditional-fp-tree_example_1}}
\end{figure*}
\begin{figure*}[!htb]
\begin{center}
\begin{tikzpicture}
\node [rectangle,draw]{ROOT} [level distance=10mm,sibling distance=25mm]
child { node [rectangle,draw]{ana=NAM (1,0)} [level distance=10mm ,sibling distance=42mm]
child {node [rectangle,draw,gray] {ant=NAM (1,0)}  [level distance=10mm ,sibling distance=27mm]
child {node [rectangle,draw,gray,dashed] {head=F (1,0)}}
child {node [rectangle,draw,gray,dashed] {head=T (0,1)}}
}
child {node [rectangle,draw,gray,dashed] {head=F (1,0)}
child {node [rectangle,draw,gray,dashed] {ant=NOM (1,0)}
}}
};
\end{tikzpicture}
\end{center}
\caption{Conditional FP-Tree for the $p=\{\text{head=F},\text{ant=NAM}\}$ pattern.\label{conditional-fp-tree_example_2}}
\end{figure*}
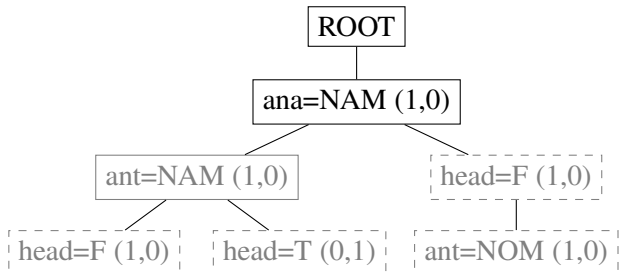

Figure~\ref{conditional-fp-tree_example_1} and Figure~\ref{conditional-fp-tree_example_2} 
show the conditional FP-Trees that are built based on the FP-Tree of Figure~\ref{final-fp-tree_example} 
and for patterns $p=\{\text{head=F}\}$ and $p=\{\text{head=F},\text{ant=NAM}\}$, respectively.
%
\subsection{Discriminative Pattern Mining vs. Feature Selection}
In this paper, we used a discriminative pattern mining approach for determining feature-values 
that are informative for the coreference label when they are considered in combination.

An alternative approach would be to use a standard feature selection algorithm where each feature-value is considered as a feature.
There are three feature selection models: \emph{filter}, \emph{wrapper} and \emph{embedded}.

\emph{Wrapper} models use a learning algorithm, i.e.\ coreference resolver in our scenario, in the loop, and therefore assess the performance
of different feature subsets based on the performance of the learning algorithm on an evaluation set.
Wrappers are, however, 
computationally expensive in our scenario since they require the coreference resolver to be executed in
every iteration of the feature-value subset selection. For $n$ feature-values, there exist $2^n$ possible combinations. $n$ is around 500 in our data and deep-coref takes two days for training using GPUs.

\emph{Filter} models, on the other hand, are solely data dependent and therefore are independent of the learning algorithm.
The use of a discriminative pattern mining approach for informative feature-value selection, is equivalent to a filter model.

Finally, embedded models are
incorporated into the learning algorithm itself.
For instance, we can incorporate all possible feature-values in deep-coref and use 
various regularization methods, e.g.\ dropouts, l$_1$ and l$_2$ regularizations, instead of pre-selecting informative feature-values.
We examined the above regularization methods in ``top-pairs+linguistic'' experiments.
The use of each of the above regularizations on top of the feature layer in deep-coref
results in significantly lower performance than either of ``top-pairs'' and ``top-pairs+linguistic'' results on the CoNLL development set.
It is worth mentioning that we did not perform hyperparameter optimization for these experiments.
We examine $0.2$, $0.3$ and $0.5$ values for dropout and $0.01$ as the regularization parameter.
If we want to tune these parameters, the question would be the choice of the evaluation set since our main focus is to improve generalization. We leave this direction for future work.

Overall, it is worth noting that EPM uses an exhaustive search to explore all frequent combination of feature-values up to a certain length, unlike many existing feature selection algorithms that use heuristic algorithms for searching feature subsets.

 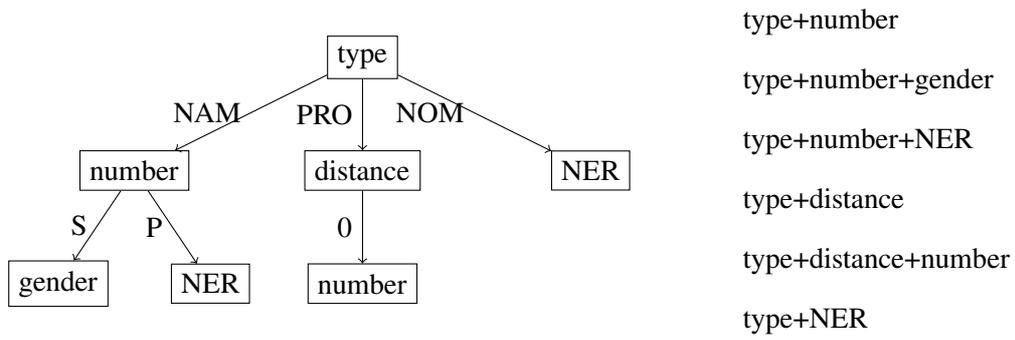
\begin{figure*}[!htb]
 \begin{minipage}{0.55\textwidth}
\begin{tikzpicture}
\node [rectangle,draw]{type} [level distance=15mm ,sibling distance=30mm]
child {node [rectangle,draw] {number}  [level distance=15mm ,sibling distance=20mm]
child {node [rectangle,draw] {gender}
edge from parent [->] node [left] {S}
}
child {node [rectangle,draw] {NER}
edge from parent [->] node [left] {P}
}
edge from parent [->] node [left] {NAM}
}
child {node [rectangle,draw] {distance}
child {node [rectangle,draw] {number}
edge from parent [->] node [left] {0}
}
edge from parent [->] node [left] {PRO}
}
child {node [rectangle,draw] {NER}
edge from parent [->] node [left] {NOM}
}
;
\end{tikzpicture}
 \end{minipage}
 \begin{minipage}{0.35\textwidth}
  \begin{itemize}
   \item[] type+number
   \item[] type+number+gender
   \item[] type+number+NER
   \item[] type+distance
   \item[] type+distance+number
   \item[] type+NER
  \end{itemize}

 \end{minipage}
\caption[A sample decision tree and its corresponding feature conjunctions]{A sample decision tree and the list of all extracted feature conjunctions based on Fernandes et al.'s (2012) approach. \label{fernandes_tree}}
\end{figure*}
\subsection{Example of Fernandes et al.'s (2012) Feature Templates}
Figure~\ref{fernandes_tree} shows a sample decision tree the list of corresponding feature templates that are extracted from it.